\documentclass[conference]{IEEEtran}
\IEEEoverridecommandlockouts

\usepackage{tabularx}
\usepackage[numbers,sort&compress]{natbib}
\usepackage{booktabs}
\usepackage{amsmath,amssymb,amsfonts}
\usepackage{algorithm}
\usepackage{algorithmic}
\usepackage{graphicx}
\usepackage{textcomp}
\usepackage{xcolor}
\usepackage[utf8]{inputenc}

\begin{document}

\title{KETM:A Knowledge-Enhanced Text Matching method\\
}
\makeatletter
\newcommand{\linebreakand}{%
  \end{@IEEEauthorhalign}
  \hfill\mbox{}\par
  \mbox{}\hfill\begin{@IEEEauthorhalign}
}
\makeatother
\author{Kexin Jiang\textsuperscript{1},Yahui Zhao\textsuperscript{1,*}, Guozhe Jin\textsuperscript{1}, Zhenguo Zhang\textsuperscript{1}, Rongyi Cui\textsuperscript{1}\\
\textsuperscript{1}Inst. of intelligent information processing,Yanbian University,Yanji 133002,China\\
\textsuperscript{*}Corresponding author\\
\texttt{email:jiangkexin1997@gmail.com,(yhzhao,jinguozhe,zgzhang,cuirongyi)@ybu.edu.cn}}


\maketitle

\begin{abstract}
Text matching is the task of matching two texts and determining the relationship between them, which has extensive applications in natural language processing tasks such as reading comprehension, and Question-Answering systems. The mainstream approach is to compute text representations or to interact with the text through attention mechanism, which is effective in text matching tasks. However, the performance of these models is insufficient for texts that require commonsense knowledge-based reasoning. To this end, in this paper, We introduce a new model for text matching called the Knowledge Enhanced Text Matching model (KETM), to enrich contextual representations with real-world common-sense knowledge from external knowledge sources to enhance our model understanding and reasoning. First, we use Wiktionary to retrieve the text word definitions as our external knowledge. Secondly, we feed text and knowledge to the text matching module to extract their feature vectors. The text matching module is used as an interaction module by integrating the encoder layer, the co-attention layer, and the aggregation layer. Specifically, the interaction process is iterated several times to obtain in-depth interaction information and extract the feature vectors of text and knowledge by multi-angle pooling. Then, we fuse text and knowledge using a gating mechanism to learn the ratio of text and knowledge fusion by a neural network that prevents noise generated by knowledge. After that, experimental validation on four datasets are carried out, and the experimental results show that our proposed model performs well on all four datasets, and the performance of our method is improved compared to the base model without adding external knowledge, which validates the effectiveness of our proposed method. The code is available at https://github.com/1094701018/KETM

\end{abstract}
\section{Introduction}
Text matching refers to taking two texts as input and determining their relationship by understanding their respective semantics, which is an important task in natural language processing\cite{jiang2022deim}. It is a fundamental technique for various tasks and has been successfully applied in many areas of natural language processing. For example, reading comprehension \cite{sugawara2020assessing}, question and answer systems \cite{liu2020asking}, and machine translation \cite{li2021incorporating}. The existing text matching models can be grouped into two main categories:  traditional methods and deep learning. The traditional text matching methods mainly rely on manually defined features to calculate the similarity between texts. It is difficult to extract deep semantic information with these methods. In recent years, due to the rapid development of deep learning and the release of related large-scale datasets with annotations, such as Snli \cite{bowman2015large} and MultiNLI \cite{williams2018broad}, deep learning-based methods are receiving increasing attention for text matching problems. The main idea is to encode two sentences into vectors by deep learning methods\cite{shen2020learning}\cite{zhou2016learning} or to interact with two sentences using attention mechanisms\cite{yang2019simple}\cite{kim2019semantic}\cite{jiang2022difm}.In order to enable the model to learn based on a better initial state and thus be able to achieve better performance. In recent years, pre-trained language models have become the dominant approach nowadays. ELMo\cite{peters2018deep} uses bi-directional LSTM to extract contextual semantic features. BERT\cite{devlin2019bert} and RoBERTa\cite{liu2019roberta} use Transformer\cite{vaswani2017attention} as the basic encoder and they achieved good results on multiple tasks in NLP. All these methods can extract sentence semantic information effectively, so their performance is higher than text matching based on traditional methods. 

The above models have achieved very good results in text matching, but the performance of these models is not perfect when the size of the data is sparse or in cases that require some common sense knowledge to judge, because these models rarely introduce external knowledge, notwithstanding the great significance it possesses in text matching. External knowledge plays a significant role in text matching, for example, adding some common sense knowledge enables the computer to infer the relationship between texts quickly, and the introduction of external knowledge can make the model achieve better results in the case of insufficient data volume. Therefore, in this paper, we propose a knowledge-enhanced text matching method abbreviated as KETM, which takes Wiktionary as the external knowledge source and adopts word paraphrases as the external knowledge. We add them to the text-matching model. Table \ref{tab:0} shows a sample case on the Snli dataset we selected, and the red part is the added external knowledge. As can be seen from Table \ref{tab:0}, without adding the knowledge, it is difficult for the model to infer that the saxophone belongs to a musical instrument, however, adding the definition of saxophone, which states that it is a musical instrument, enables the model to reason correctly about their relationship. We conduct experiments on several datasets and verify the effectiveness of the method on the text matching task. Our contributions can be summarized as follows:
\begin{itemize}
\item[$\bullet$] We propose a knowledge-enhanced text matching method——KETM, which adds the interpretation of words to the model and fuses them with the text information using a fusion of gating mechanisms, With this method, the model will be able to focus on the  semantic information of text pairs while also pouncing on the semantic information of the words in them and effectively fusing the information of text and knowledge.
\end{itemize}
\begin{itemize}
\item[$\bullet$] Our proposed method has good generality and can be applied to other text-matching models to improve the performance of the model without adding too many additional model parameters.
\end{itemize}
\begin{itemize}
\item[$\bullet$]  We experiment on several well-known datasets. Our proposed text-matching model achieves very good performance, and the experimental results are all improved compared to the baseline model on these datasets. 
\end{itemize}
\begin{table*}
\centering
  \caption{\label{tab:0}Snli Premise (P) and Hypothesis (H) and Word Definition (red) from Wiktionary.}
\begin{tabular}{c}
\begin{tabular}{c}
 \toprule P: The man is holding a \color{red}{saxophone.}\\\color{red}{`saxophone': ``A single-reed instrument musical instrument of the woodwind}\\ \color{red}{family, usually made of brass and with a distinctive loop bringing the bell upwards.''} \\
 \midrule
  \end{tabular}
\\
H:The man is holding an \color{red}{instrument}\\\color{red}{
 `instrument': ``A device used to produce music.''}\\
\bottomrule
\end{tabular}
\end{table*}

\section{Related Work}

The deep learning-based text matching models are divided into two categories; (1)Representation-based text matching models, the main purpose of which is to represent a sentence as a vector. (2) Interaction-based text matching models, which aim to obtain complex interaction information between sentences.

\subsection{Representation-based text matching model}

Representation-based text matching models focus on constructing a representation vector of sentences. The traditional methods for text matching include the following: similarity-based methods \cite{renhan2015recognizing}, rule-based methods \cite{hu2020extended}, alignment feature-based machine learning methods \cite{sultan2015feature}, etc. In recent years, deep learning-based approaches have been significantly effective in semantic modeling and have achieved good results in many tasks in NLP. Therefore, on the task of text matching, the performance of deep learning-based methods has surpassed the earlier methods and has become the mainstream text matching method. For example, Bowman {\it et al}.\cite{bowman2015large}  first applied LSTM sentence models to the task by encoding premises and hypotheses through LSTM to obtain sentence vectors. Facebook proposed the Infersent\cite{conneau2017Supervised} model for sentence embedding, and the authors came to the conclusion that the best results were obtained using BiLSTM combined with maximum pooling by comparing several CNN and RNN-based encoders. Reimers {\it et al}.\cite{reimers2019sentence} proposed the SBERT model based on its two-tower structure. The model uses BERT for sentence encoding to obtain the sentence vectors. The conclusion shows a substantial improvement by using the fusion strategy over the direct splicing of the output using BERT. Gao {\it et al}.\cite{gao2021simcse}  proposed the SimCSE model, which also solves the sentence embedding problem. It uses BERT to obtain sentence embeddings and uses the idea of contrast learning to enlarge the spacing of unrelated samples.

\subsection{Interaction-based text matching model}

The main focus of the two-tower model is on the optimization of the encoder. The core idea of the interactive model is two-by-two interaction, that is, all words in two text sequences interact with information one by one. Most of the current choices are attention mechanisms, and the efficiency of the two-two interaction will decrease while the accuracy rate increases. It is possible to add encoders before and after the interaction, and then let the vectors be spliced, differenced, and dotted to improve the effect.

PARIKH  {\it et al}.\cite{parikh2016decomposable} proposed a lightweight interaction model DecAtt, which allows individual words in a sequence of sentences to be compared with each other to obtain word-level synonymy and antonymy. The computational lightweight has better results at the same time, but it is prone to the problem of excessive gradient if the sequence is too long. He {\it et al.}.\cite{he2016pairwise} proposed the fine-grained interaction-oriented model PWIM, which used BiLSTM to complete the encoding of input sequences with a deep model structure and proposed functions consisting of cosine, dot product, and Euclidean distance for vector comparison, but the overall computational complexity of the model was too large and was overtaken by the later proposed ESIM.\cite{chen2017enhanced}. ESIM model that uses a two-layer bidirectional LSTM and a self-attentive mechanism for coding, and then extracting features through the average pooling layer and the maximum pooling layer, and finally performs classification. Meanwhile, Wang {\it et al}.\cite{wang2017bilateral} proposed the BIMPM model, which first encodes sentence pairs through a bidirectional LSTM, and then matches encoding results from multiple perspectives in two directions. Compared to ESIM with larger dimensional vectors, Tay {\it et al}.\cite{tay2018compare} proposed the CAFÉ model, which compressed the feature vectors and achieved a lightweight model. Previous models have mostly adopted interaction, but the number of interactions is small. Therefore, Yang {\it et al}.\cite{yang2019simple} proposed a simple and efficient text matching model, The model has a significant improvement based on multiple interactions with the most original encoding vector spelled out each time.



\begin{figure*}[h]
\centering
\includegraphics[width=4.8in]{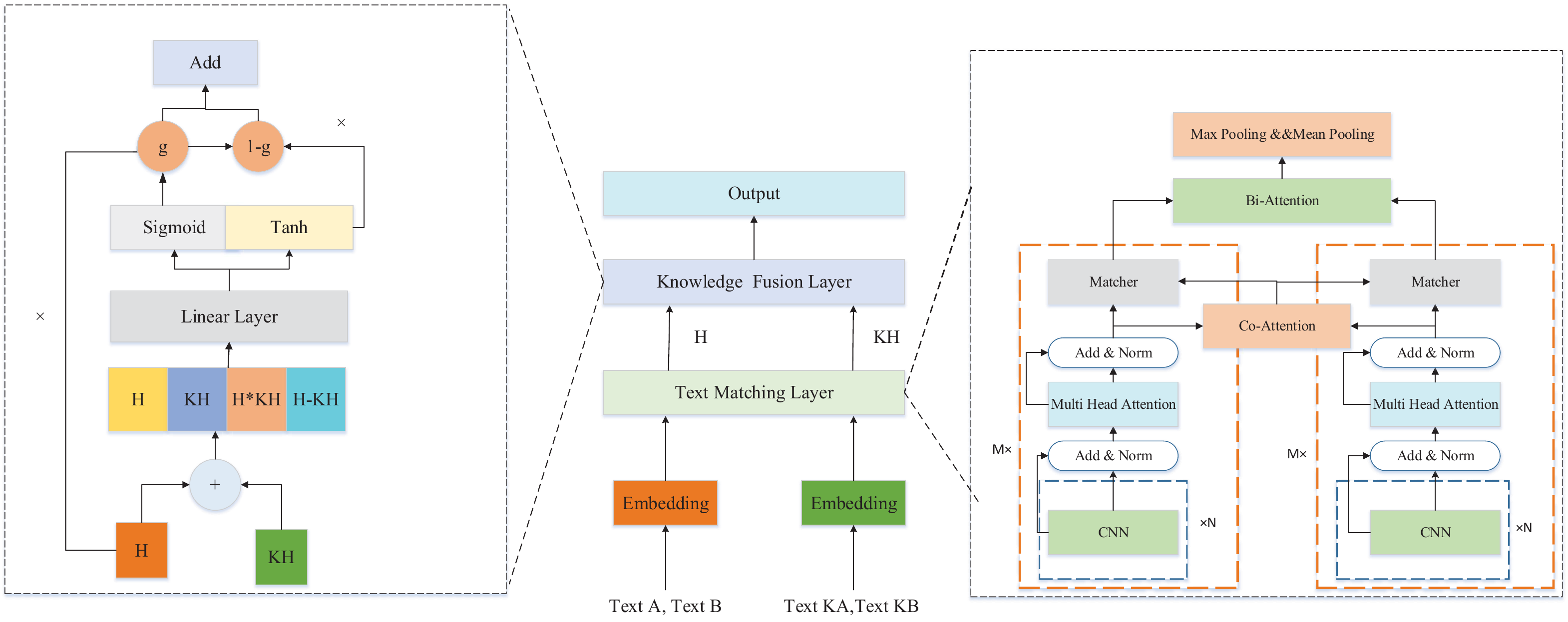}
\caption{Overview of the architecture of our proposed KETM model. The middle shows the overall model framework proposed in this paper, the left side shows the flow chart of the knowledge fusion layer, and the right side shows the text matching layer proposed in this paper, which can be an arbitrary text matching model}
\label{fig:1}
\end{figure*}
\section{Method}
\subsection{knowledge retrieval}
In this paper, we choose Wiktionary\footnote{https://www.wiktionary.org/}, an online dictionary, to be the external knowledge source. Wiktionary contains 999,614 entity descriptions, and for each entity of our text, we choose its first entity description in Wiktionary as our external knowledge. We looked up the closest match of each word in Wiktionary by using the lemma
form by Spacy. For example, the word
“singing” does not appear in its original form
in Wiktionary, but its lemma form “sing” is
in Wiktionary and we get its description text: “To produce musical or harmonious sounds with one’s voice”. In this way, we find descriptions
of all entities in our experiments. We assume that the input sentence pairs are $A$ and $B$. We retrieve the paraphrases of the words in Wiktionary and splice them together. We get the knowledge text $KA$, and $KB$ respectively.
\subsection{Model}
In this subsection, we will describe our model in detail. As shown in Figure~\ref{fig:1}, our model  mainly consists of an embedding layer, a text matching layer, a knowledge fusion layer, and an output layer.
\subsubsection{Embedding Layer}
For the above texts $A$, $B$ and for the knowledge texts $KA$, $KB$, we use the pre-trained language model F for word embedding, the process is shown in Eq.\ref{eq1}:
\begin{equation}\label{eq1}
\begin{aligned}
& X=F\left( A \right) \in R^{m*k} \\
& Y=F\left( B \right) \in R^{n*k} \\
\end{aligned}
\end{equation}
where $F$ is denoted as a pre-trained language model, and in this paper, we choose ELMo and BERT\_base. $m$,$n$ are the number of words of text $A$ and text $B$, respectively, and $k$ is the hidden layer dimension. When $F$ is ELMo, the value of $k$ is 1024, and when $F$ is BERT, $k$ is 768.

We perform the same process on the knowledge text to obtain its word vector $KX$,$KY$.
\subsubsection{text matching Layer}
In our model framework, the layer can be any text-matching model $M$. Its output can be defined formally as $H=M(X,Y)\in R^{d}$, where $d$ is the hidden layer dimension. Similarly, for knowledge text, there is $KH=M(KX,KY)\in R^{d}$.

In this section, we propose an effective text matching model for deep encoding and interaction, which includes an encoding layer, a cross-attention layer, an aggregation layer, a bidirectional attention layer, and a pooling layer
\paragraph{\textbf{encoding layer}}
The purpose of the  encoder layer is to fully exploit the contextual relationship features of the sentences. Generally, most existing models use bidirectional LSTM for encoding\cite{wang2017bilateral}\cite{chen2017enhanced}, which can extract the contextual relationship features of the sentence, but there still exists the problem of gradient disappearance for long sentence sequences. Meanwhile, since the LSTM adopts the architecture of RNN, 
it may affect the efficiency of the encoding process. Therefore, we use CNN and multi-head attention for encoding. Local features are first extracted using multiple CNNs in the encoding layer, and then global features are obtained using multi-head attention. We take text A as an example, and its formula is shown in Eq.\ref{eq2}:
\begin{equation}\label{eq2}
\begin{aligned}
  &  A_c=Conv\left( X \right) 
\\
& A_m=MultiHead\left( \left[ A_c:X \right] \right) 
\\
& P=\left[ A_c:A_m \right] \in R^{m*d}
\end{aligned}
\end{equation}
where $A_c$ and $A_m$ are the output of the convolution layer and multi-head attention layer, respectively; [$A_c$:$X$] and [$A_c$:$A_m$] represent the concatenation operation. In the same way, for text B, we obtain the output matrix,
namely $H \in R^{n*d}$.
\paragraph{\textbf{co-attention layer}}
After obtaining the encoding $P$ and $H$ of text $A$ as well as text $B$, we use cross-attention for the first interaction. This is done as follows: First, the similarity matrix $S$ between $P$ and $H$ is calculated, then it is normalized to obtain the attention weights, and finally, it is weighted and averaged to obtain the text representations $X$ and $Y$, the process is shown in Eq.\ref{eq3}:
\begin{equation}\label{eq3}
\begin{aligned}
    & S=relu{{({{W}_{c}}{{P}^{T}})}^{T}}relu({{W}_{q}}{{H}^{T}}) \\ 
 & a=softmax(S) \\ 
 & {{P}^{'}}=a\bullet H \\ 
 & {{H}^{'}}={{a}^{T}}\bullet P \\ 
\end{aligned}
\end{equation}
where ${W_c}$ and ${W_q}$  are the learnable parameters. 
\paragraph{\textbf{aggregation layer}}
The aggregation layer is the aggregation of the three perspectives before and after cross-attention, and we take text A as an example, the process is shown in Eq.\ref{eq4}:
\begin{equation}\label{eq4}
\begin{aligned}
& {a}_{1}={G}_{1}\left( \left[ {P};{P}^{'} \right] \right)  \\ 
 & {a}_{2}={G}_{2}\left( \left[ {P};{P}-{P}^{'} \right] \right)   \\ 
 & {a}_{3}={G}_{3}\left( \left[ {P};{P}\odot {P}^{'} \right] \right) \\ 
 & {{C}={G}\left( \left[ {a_1};{a_2};{a_3} \right] \right)}  \\
\end{aligned}
\end{equation}
where $G_1$,$G_2$,$G_3$,$G$ is a forward neural network, $\odot$ is the element-level multiplication, the subtraction operator reflects the difference between two vectors, and the multiplication operator reflects the similarity of two vectors. Similarly, the representation $Q$ of text $B$ can be obtained.
\paragraph{\textbf{Bidirectional Attention Layer}}
The purpose of the bidirectional attention layer is to better integrate the characteristics of the two aspects. We calculate the bidirectional attention of $C$ and $Q$ that is the attention of $C\to Q$ and $Q\to C$ \cite{seo2016bidirectional}. The attention  originates from the similarity matrix {$T$}, where ${{t}_{ij}}$ denotes the similarity between the $i$-th word of $C$ and the $j$-th word of $Q$.

$C\to Q$: The attention  describes which words in the text $C$ are most relevant to $Q$.The calculation process is similar to that of the co-attention layer, We can obtain the attention matrix $U\in {{R}^{d*n}}$, which is calculated as shown in Eq.\ref{eq5}.
\begin{equation}\label{eq5}
\begin{aligned}
   & {{\alpha }_{t}}={softmax} ({{T}_{t:}})\in {{R}^{n}} \\ 
 & {{u}_{:t}}=\sum\limits_{j}{{{\alpha }_{tj}}{{Q}_{:j}}}  
\end{aligned}
\end{equation}
where ${T_{t:}}$ is the $t$-th raw of $T$, and ${u_{:t}}$ is the $t$-th column of $U$.

$Q\to C$: The attention indicates which words in $Q$ are most similar to $C$. The calculation process is as follows: first, the column with the largest value in the similarity matrix {$T$} is taken to obtain the attention weight, then the weighted sum of $C$ is expanded by $n$ time steps to obtain $V\in {{R}^{d*n}}$, which is calculated as shown in Eq.\ref{eq6}.
\begin{equation}\label{eq6}
\begin{aligned}
    & b=softmax (\underset{col}{\mathop{\max }}\,(T))\in {{R}^{m}} \\ 
 & v=\sum\limits_{t}{{{b}_{t}}{{H}_{t:}}\in {{R}^{d}}} \\ 
\end{aligned}
\end{equation}

After we obtain $U$ and $V$, we stitch the attention in these two directions by a multilayer perceptron to obtain the contextual representation $G\in {{R}^{4d*n}}$, which is calculated as shown in Eq.\ref{eq7}.
\begin{equation}\label{eq7}
\begin{aligned}
    & {{G}_{:t}}=\beta ({{V}_{:t}},{{C}_{:t}},{{U}_{:t}}) \\ 
 & \beta (v,c,u)=[c;v;c\odot v;c\odot u]\in {{R}^{4d}}  
\end{aligned}
\end{equation}
\paragraph{\textbf{Pooling Layer}}
The purpose of the pooling layer is to extract the key information of the text. In this paper, average pooling and maximum pooling are used. Their outputs are directly spliced, which is calculated as shown in Eq.\ref{eq8}.
\begin{equation}\label{eq8}
\begin{aligned}
    & G_{\max}=Max\left( G \right)\\ 
   & G_{\mathrm{m}ean}=Mean \left( G \right)\\  
   & H=\left[ G_{\max};G_{\mathrm{m}ean} \right]\\
\end{aligned}
\end{equation}
where $Max$ denotes maximum pooling, $Mean$ denotes average pooling, and $H$ denotes the pooled vector
\subsection{ Knowledge Fusion Layer}
After the text matching model $M$, we obtain the text representation $H$ as well as the knowledge representation $KH$. To reduce the influence of noise introduced by the fusion module, we use a gating mechanism in our fusion layer. We use a neural network to control the fusion ratio of text and knowledge vectors. We fuse $H$ and $KH$ to obtain the text representation $Z=fusion(H,KH)$ , where the fusion function is defined as shown in Eq.\ref{eq9}:
\begin{equation}\label{eq9}
\begin{aligned}
    & \widetilde{x}=\tanh ({{W}_{1}}[H;KH;H\odot KH;H-KH]) \\ 
 & g=sigmoid({{W}_{2}}[H;KH;H\odot KH;H-KH]) \\ 
 & z=g\odot \widetilde{x}+(1-g)\odot x \\
\end{aligned}
\end{equation}
Where ${{W}_{1}}$ and ${{W}_{2}}$  are weight matrices, and $g$ is a gating mechanism to control the weight of the intermediate vectors in the output vector.
\subsection{ Output Layer}
The purpose of the output layer is to output the results. In this paper, we use a linear layer to get the results of text matching. The process is shown  in Eq.\ref{eq10}.
\begin{equation}\label{eq10}
\begin{aligned}
   y = softmax(\tanh (ZW + b))
\end{aligned}
\end{equation}
where both $W$ and $b$ are trainable parameters.

Finally, we use the cross-entropy function to calculate the loss, the cross-entropy loss function can be chosen as shown  in Eq.\ref{eq11}.
\begin{equation}\label{eq11}
\begin{aligned}
   loss=-\sum\limits_{i=1}^{N}{\sum\limits_{k=1}^{K}{{{y}^{(i,k)}}\log {{{\hat{y}}}^{(i,k)}}}}
\end{aligned}
\end{equation}
where $N$ is the number of samples, $K$ is the total number of categories, and ${{\hat{y}}^{(i,k)}}$ is the true label of the $i$-th sample. 
\section{Experiments}
In this section, we first present some details of the experiment implementation, and secondly, we show the experimental results on the dataset. Finally, we analyze and discuss the experimental results.
\subsection{Experimental details}

\subsubsection{Dataset}
In this paper, we use the text-matching datasets Snli, SciTail,  Quora, and Sick to validate our model. Among them, the Snli dataset includes 570K manually labeled and categorically balanced premise and hypothesis pairs. The SciTail dataset includes 27k pairs of premise and hypothesis that contain an entailed or neutral relationship. The Quora question pair dataset includes over 400k pairs of data that each with binary annotations, with 1 being a duplicate and 0 being a non-duplicate. The Sick dataset contains 10K sentence pairs involving knowledge of morphology, including implication, neutrality, and contradiction labels. The statistical descriptions of Snli, SciTail, Quora, and Sick data are shown in Table \ref{tab:1}.
\begin{table}[htbp]
 \centering
  \caption{\label{tab:1}The statistical descriptions of Snli, Scitail, Quora, and SICK}
\begin{tabular}{cccc}
 \toprule
dataset & train & validation  & test\\
 \midrule
Snli & 550152 & 10000 & 10000 \\
SciTail & 23596 & 1304 & 2126 \\
Quora & 384290 & 10000 & 10000 \\
Sick & 4500 & 500 & 4927 \\
\bottomrule
 \end{tabular}
\end{table}
\subsubsection{Baseline methods and parameter settings}
We compare nine baseline methods on the dataset, including  representation-based models (i.e., SWEM (Shen et al.2018), HBMP (Talman et al.2018)), and interaction-based models (i.e., RE2 (Yang et al.2019), DITM(Yu et al.2021), DRr-Net(Zhang et al.2019), BIMPM(Wang et al.2017), ESIM(Chen et al.2017)), and pre-training based models (i.e., BERT(Devlin et al.2019), MFAE(Zhang et al.2020)).

This experiment is conducted in a hardware environment with a graphics card RTX5000 and 16G of video memory. The system is Ubuntu 20.04, with the development language Python 3.7 and the deep learning framework Pytorch 1.8.

In the model training process,  the maximum length of the text is set to 128, and the hidden layer dimension is set to 200. The specific hyperparameter settings are shown in Table \ref{tab:2}.
\begin{table}[htbp]
 \centering
  \caption{\label{tab:2}Values of Hyper Parameters}
\begin{tabular}{cc}
 \toprule
Hyper Parameters & Values\\
 \midrule
hidden dimension & 200 \\
convolution kernel size & 3 \\
learning rate & 0.00005 \\
Optimizer & Adam \\
Dropout & 0.2 \\
activation function & ReLU \\
Epoch & 30 \\
Batch size & 48 \\
\bottomrule
 \end{tabular}
\end{table}
\subsection{Experimental results}
We compare the experimental results of  our model on these datasets with other published models. The evaluation metric we use is the accuracy rate. The results are shown in Table \ref{tab:3}. where KETM denotes we use ELMo for word embedding. KETM(BERT) denotes word embedding using BERT and splicing cls vectors after the text matching layer. KETM-KB denotes that our model does not use external knowledge, indicating that the model has only the embedding layer and the text-matching layer. * denotes our implementation results. 

As can be seen from Table \ref{tab:3}, our model achieves  90.6\%, 92.6\%, 91.0\%, and 87.1\% accuracy on the Snli, Scitail, Quora, and Sick test sets, respectively, which is best among the listed methods. Overall, the performance of the representation-based text matching model is slightly weaker than that of the interaction-based model, And the slight weakness may result from the ignorance of the complex interaction information between texts of the representation-based approach, which only focuses on the information of the text itself.

Compared to the representation-based matching models, our models all show substantial improvements in results. On the three datasets (Snli, Scitail, and Quora), the accuracy is improved by 4.0, 6.6, and 7.0 percentage points, respectively, over the best-performing model. Compared to the interaction-based model, our model improves the accuracy by 1.9, 6.0, 1.6, and 7.3 percentage points over the higher-performing RE2 model on the four datasets, respectively. It is the best among the listed methods. Compared to the pre-trained language model BERT\_base and the MFAE model that use BERT as an encoder. Our model is also the best on the four datasets with 0.6, 0.6, 0.5, and 1.7 percentage points higher, respectively.

Compared to models without external knowledge, the accuracy of our model improves by 0.6, 0.9, 0.2, and 3.9 percentage points on the four datasets when we use ELMo as the word vector, respectively. The improvement on the Snli dataset and Quora dataset is not significant, probably because their data volume is already large and the accuracy rate is higher without adding knowledge, and adding external knowledge won't have much big effect. However, the improvement result on the Sick dataset is significant. It shows that the addition of external knowledge has a positive effect on the overall performance of the model. We conduct experiments from multiple perspectives, and the experimental results verify the effectiveness of our model
\begin{table}[htbp]
 \centering
  \caption{\label{tab:3}The accuracy($\%$) of the model on the Snli, Scitail, Quora, and Sick test sets. Bold black is the best performance of the listed methods}
\begin{tabular}{ccccc}
 \toprule
Model & Snli & Scitail & Quora & Sick \\
 \midrule
SWEM\cite{shen2018baseline} & 83.8  & -  & 83.0 & -\\
HBMP\cite{talman2018natural} & 86.6  & 86.0  & - & -\\
 \midrule
DITM\cite{yu2021simple} & - & 89.2 & 86.1 & -\\
DRr-Net\cite{zhang2019drr} & 87.7 & 87.4 & 89.8 & -\\
BIMPM$^*$\cite{wang2017bilateral} & 87.9 & 75.3 & 88.2 &76.6 \\
ESIM$^*$\cite{chen2017enhanced} & 88.0 & 82.4 & 85.4 & 76.4 \\
RE2$^*$\cite{yang2019simple} & 88.7 & 86.6  & 89.4 & 79.8 \\ 
 \midrule
MFAE\cite{zhang2020questions} & 90.0 & - & 90.5 & - \\
BERT\_base$^*$\cite{devlin2019bert} & 89.6 & 92.0 & 89.9 &85.4 \\
 \midrule
KETM-KB$^*$ & 88.9 & 89.5 & 90.1 & 80.1 \\
KETM$^*$ & 89.5 & 90.4  & 90.3 & 84.0 \\ 
KETM-KB(BERT)$^*$ & 90.2 & 92.1 & 90.7 & 86.8 \\
\textbf{KETM(BERT)}$^*$ & \textbf{90.6} & \textbf{92.6}  & \textbf{91.0} & \textbf{87.1} \\ 
\bottomrule
 \end{tabular}
\end{table}
\subsection{ Analysis of the method generality}
To verify the generality of the knowledge enhancement method proposed in this paper, we conduct experiments using the text matching model ESIM and the pre-trained language model BERT. The experimental results are shown in Figure \ref{fig:2}.
\begin{figure}[h]
\centering
\includegraphics[width=3.6in]{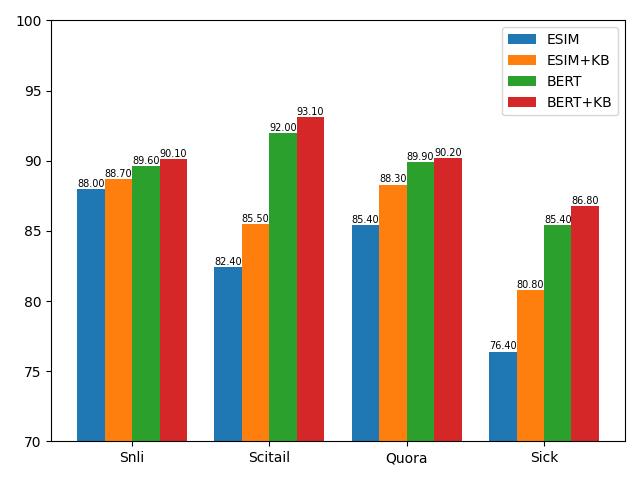}
\caption{The accuracy($\%$) of the different models on the Snli, Scitail, Quora, and Sick test sets.}
\label{fig:2}
\end{figure}
As can be seen from Figure \ref{fig:2}, on the non-transformer architecture ESIM model, we add external knowledge and improve the accuracy of the four datasets by 0.7, 3.1, 2.9, and 4.4 percentage points, respectively. Meanwhile, we add external knowledge on the Transformer architecture BERT model, and the accuracy of the model improves by 0.5, 1.1, 0.3, and 1.4 percentage points, respectively, and the performance is improved on all of them. We achieve better results on both the non-Transformer architecture-based and Transformer architecture-based models compared with no knowledge addition, which validates the effectiveness of our proposed knowledge enhancement framework.
\subsection{ Analysis of the different training data}
To verify the role of knowledge in different training data sizes, we selected part of the training set of the Snli dataset for training. The experimental results are shown in Figure \ref{fig:3}.
\begin{figure}[h]
\centering
\includegraphics[width=3.6in]{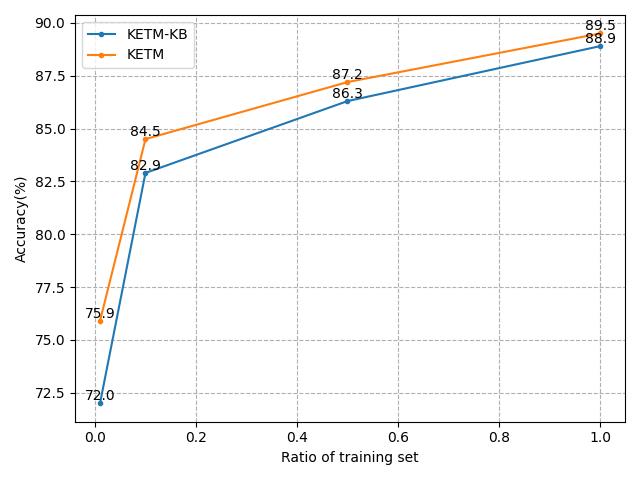}
\caption{Experimental results for different training set sizes on the Snli dataset.}
\label{fig:3}
\end{figure}
As can be seen from Figure \ref{fig:3}, when we train with 1\% of training data, the model improves the most performance when external knowledge is added, indicating that external knowledge achieves a larger role with a small amount of data. Meanwhile, when we train with 10\%, 50\%, and 100\% of the training data, the accuracy is improved by 1.6, 0.9, and 0.6 percentage points after adding knowledge, respectively. The experimental results show that external knowledge plays a facilitating role in different training data sizes. Besides, the smaller the training data is, the greater improvement the model achieves.
\subsection{ Analysis for the adversarial dataset}
To verify the knowledge of the adversarial dataset, we train on the Snli training set and test on the adversarial dataset BreakNLI dataset\cite{glockner2018breaking}. The BreakNLI dataset is mainly designed to test the model's inferential knowledge of lexical words. The premise of this test set is composed of sentences from the Snli training set, and the sentences are assumed to replace one of the words to obtain new text pairs. Implicative samples are generated by replacing words with their synonyms or superlatives; contradictory samples are generated by replacing words with their mutually exclusive words, and neutral samples are generated by replacing words with their subordinates. The experimental results are shown in Table \ref{tab:7}.
\begin{table}
 \centering
  \caption{\label{tab:7}The accuracy($\%$) of the model on the adversarial dataset BreakNLI. }
 \begin{tabular}{cc}
\toprule
Model & BreakNLI \\
\midrule
BIMPM$^*$ & 68.3\\
RE2$^*$ & 80.9\\
KIM & 83.8\\
\midrule
ESIM & 65.8\\
ESIM+KB$^*$ & 78.8\\
\midrule
KETM-KB$^*$ & 87.7 \\
\textbf{KETM$^*$} & \textbf{91.2} \\
\bottomrule
 \end{tabular}
\end{table}
As can be seen from Table \ref{tab:7}, the models ESIM, BIMPM, and RE2, which originally performed well on the Snli dataset, showed a significant decrease in effectiveness on the BreakNLI dataset. The accuracy of the ESIM model is 13 percentage points higher when external knowledge is added to the ESIM model. Also compared to the model proposed in this paper, adding external knowledge improves 3.5 percentage points on the adversarial dataset. The main reason for this may be that our method incorporates word interpretation, so the model can determine the relationship between words based on word interpretation, and thus performs well on the adversarial dataset BreakNLI. The experimental results demonstrate the effectiveness of our model to incorporate external knowledge on the adversarial dataset.
\subsection{ Ablation experiments}
In order to verify the effectiveness of the model fusion layer as well as the text matching layer module, we conducted ablation experiments on the validation sets of sick and Scitail. Not using the fusion function means that the textual information is directly spliced with the knowledge information. The experimental results are shown in Table \ref{tab:8}.
\begin{table}
 \centering
  \caption{\label{tab:8}Ablation study  on  Sick and Scitail validation dataset. }
 \begin{tabular}{ccc}
\toprule
 & Sick & Scitail \\
\midrule
full model & 87.0 & 91.17\\
w/o fusion & 85.4 & 90.10\\
w/o bi-att & 86.0 & 90.64\\
\bottomrule
 \end{tabular}
\end{table}

\subsection{ Case study}
We randomly pick a few samples to add to the model for prediction, when we input the premise is `The blonds girl is surfing' and the Hypothetical is `A blond girl is riding the waves', the results are shown in Table \ref{tab:9}. As can be seen from Table \ref{tab:9}, we add external knowledge to correctly infer the entailment relationship between the two texts. We visualize the results and the visible result is shown in Figure~\ref{fig:4}.
\begin{figure}[htbp]
    \begin{minipage}[t]{0.5\linewidth}
        \centering
        \includegraphics[width=\textwidth]{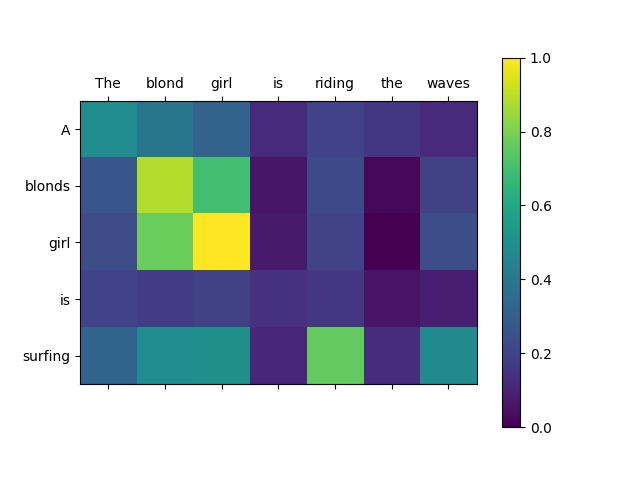}
        \centerline{(a) without external knowledge}
    \end{minipage}%
    \begin{minipage}[t]{0.5\linewidth}
        \centering
        \includegraphics[width=\textwidth]{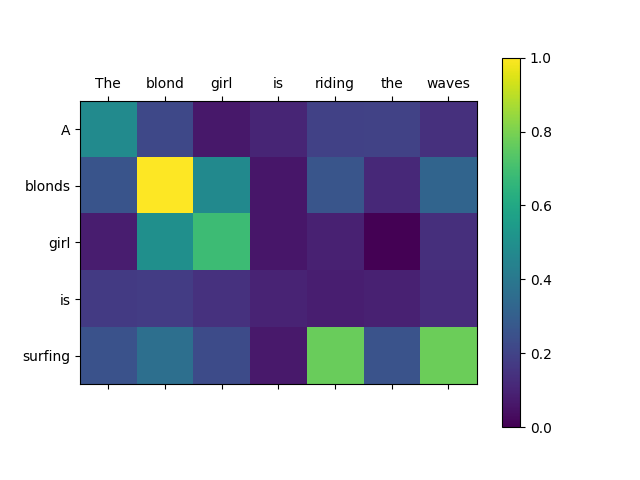}
        \centerline{(b)  add external knowledge}
    \end{minipage}
    \caption{Visualized attention weight diagram, vertical and horizontal axes represent premises and hypotheses respectively, colors indicate their weights, no external knowledge added to the left, external knowledge added to the right}
\label{fig:4}
\end{figure}
As can be seen from Figure~\ref{fig:4},
on the left is the heat map of attention without adding knowledge, and on the right is the heat map of attention adding external knowledge. When no external knowledge is added, surfing and riding waves share very few similarities, so the model decides that they are neutral. However, when external knowledge is added, the interpretation of external knowledge surfing has the meaning of surfing on a surfboard, which can link the riding waves and increase the similarity between them, therefore the model can correctly infer that the two are entailed. It shows that the introduction of external knowledge can make the model predict correctly to some extent.
\section{Conclusion}
In this paper, we investigate text matching methods and propose a text matching method based on knowledge enhancement. We enrich our model by adding the paraphrases of words to the model and use a gating mechanism to fuse the knowledge text with the original text to avoid the noise generated by knowledge. Also, the method in this paper has good generality and can function as any text-matching model in the text matching layer. We can get from the experimental results that the knowledge we added is beneficial to improve the performance of the model, and we can get from the ablation experiments that in terms of fusing knowledge, using the gating mechanism can effectively reduce the noise generated by external knowledge. 
\section*{Acknowledgements} 
This work is supported by Major program of the National Social Science Foundation of China [grant numbers 22ZD305] ,National Natural Science Foundation of China [grant numbers 62162062]. State Language Commission of China under Grant No.YB135-76. Scientific research project for building world top discipline of Foreign Languages and Literatures of Yanbian University under Grant No. 18YLPY13. The school-enterprise cooperation project of Yanbian University [2020-15].

%
%
%
%

\bibliographystyle{IEEEtran}
\footnotesize
\bibliography{ref}{}
\end{document}